\documentclass[letterpaper]{article} 
\usepackage{aaai2026}  

\usepackage{times}  
\usepackage{helvet}  
\usepackage{courier}  
\usepackage[hyphens]{url}  
\usepackage{graphicx} 
\usepackage{natbib}  
\usepackage{caption} 
\usepackage{algorithm}
\usepackage{algorithmic}

\usepackage{newfloat}
\usepackage{listings}
\DeclareCaptionStyle{ruled}{labelfont=normalfont,labelsep=colon,strut=off} 
\floatstyle{ruled}
\newfloat{listing}{tb}{lst}{}
\floatname{listing}{Listing}
\usepackage{amsmath}
\usepackage{enumitem}
\usepackage{listings}
\usepackage{multirow}
\usepackage{enumitem}
\usepackage{adjustbox}
\usepackage{todonotes}
\usepackage{booktabs}
\usepackage{caption}
\usepackage{subcaption}
\usepackage{float}
\usepackage{svg}
\usepackage{dblfloatfix}

\usepackage[capitalize]{cleveref}
\UseRawInputEncoding

\definecolor{codegreen}{rgb}{0,0.6,0}
\definecolor{codegray}{rgb}{0.5,0.5,0.5}
\definecolor{codepurple}{rgb}{0.58,0,0.82}
\definecolor{backcolour}{rgb}{0.95,0.95,0.92}
\usepackage{xcolor} 

\usepackage[most]{tcolorbox}
\usepackage[table]{xcolor}

\newtcolorbox{highlightA}{colback=green!20, colframe=yellow!50!black, boxrule=0pt, sharp corners}
\newtcolorbox{highlightB}{colback=red!15, colframe=green!50!black, boxrule=0pt, sharp corners}

\nocopyright 

\title{Who Sees the Risk?\\ Stakeholder Conflicts and Explanatory Policies in \\ LLM-based Risk Assessment}
\author{
    Srishti Yadav  \textsuperscript{\rm 1}\thanks{Work done as part of internship at IBM Research, Ireland},
    Jasmina Gajcin \textsuperscript{\rm 2},
    Erik Miehling  \textsuperscript{\rm 2},
    Elizabeth Daly \textsuperscript{\rm 2}
}
\affiliations{
    \textsuperscript{\rm 1}University of Copenhagen, Denmark\\
    \textsuperscript{\rm 2}IBM Research, Ireland\\
    srya@di.ku.dk, jasmina.gajcin2@ibm.com, erik.miehling@ibm.com, elizabeth.daly@ie.ibm.com 
    
}

\begin{document}

\maketitle

\begin{abstract}
Understanding how different stakeholders perceive risks in AI systems is essential for their responsible deployment. This paper presents a framework for stakeholder-grounded risk assessment by using LLMs, acting as judges to predict and explain risks. Using the Risk Atlas Nexus and GloVE explanation method, our framework generates stakeholder-specific, interpretable policies that shows how different stakeholders agree or disagree about the same risks. We demonstrate our method using three real-world AI use cases of medical AI, autonomous vehicles, and fraud detection domain. We further propose an interactive visualization that reveals how and why conflicts emerge across stakeholder perspectives, enhancing transparency in conflict reasoning. Our results show that stakeholder perspectives significantly influence risk perception and conflict patterns. Our work emphasizes the importance of these stakeholder-aware explanations needed to make LLM-based evaluations more transparent, interpretable, and aligned with human-centered AI governance goals.
\end{abstract}


\section{Introduction}





In recent years, unprecedented deployment of large language models (LLMs) has raised concerns about reliability, explainability and safety of these models in real world usecases. The need for responsible use of these models has led to increased interest in governance of AI models which encompasses standardized principles and evaluations to ensure that the AI systems behave reliably, robustly and reflect societal values. In order to address safety concerns, the research community has developed specialized safety benchmarks like SafetyBench \cite{zhang2023safetybench}, HarmBench \cite{mazeika2024harmbench}, SG-Bench \cite{mou2024sg} to evaluate models for harmful behaviors such as toxicity, hallucination, or misuse.




\begin{figure*}
  \centering
  \includegraphics[width=0.8\textwidth,trim={0 200 0 110},clip]{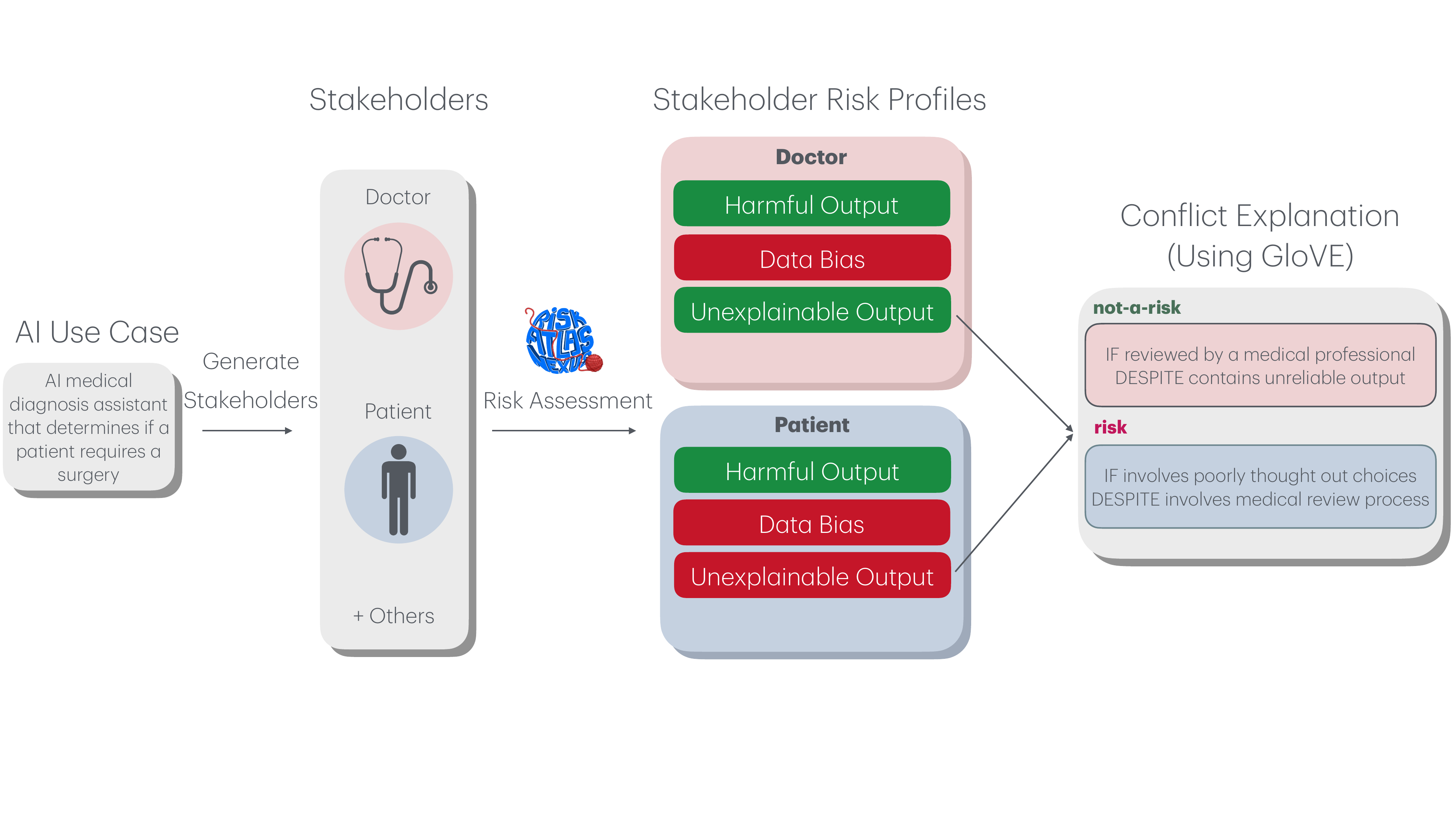}
  \caption{Overview of our stakeholder-centered AI risk assessment pipeline. The use case (``AI medical diagnosis assistant that determines if someone needs surgery") generates relevant stakeholders such as doctors, patients etc.. Each stakeholder undergoes a risk assessment that produces individual risk profiles (e.g., harmful output, data bias, unexplainable output). The GloVE component then generates conflict explanations, showing the conflicts that emerge between stakeholders’ risk perspectives }
  \label{fig:pipeline}
\end{figure*}

In order to bring structure and guidance when considering risks, several taxonomies and frameworks have emerged such as the Top 10 for LLMs and Generative AI Apps \cite{owasp}, the NIST AI Risk Management Framework \cite{nist}, the MIT AI Risk Repository \cite{mit-risk-repository, slattery2024ai}, 
and  IBM Risk Atlas Nexus \cite{bagehorn2025ai}. These risk frameworks can help link risks to inform governance mechanisms. Recent works have explored the use of LLMs take into account the use case context in order to prioritise which risks are most related to the AI system \cite{eiras2025know, Mylius2025AIGovernance, daly2025usage}. However, most existing approaches remain stakeholder-agnostic, overlooking the nuanced ways in which different stakeholders may perceive and prioritize risks which is an essential concern for responsible governance. This gap limits the effectiveness of governance tools in multi-stakeholder environments.


We propose leveraging stakeholder specific personas as part of the risk prioritisation process. By considering our framework on stakeholder perspectives, our approach reveals points of alignment and disagreement on risk assessment across stakeholders for a given AI use case. This enables more context-sensitive governance decisions and supports inclusive risk mitigation planning. We generate explanations for stakeholder conflicts using the GloVE pipeline \cite{gajcin2025interpreting} that allows for a more transparent and context-sensitive interpretation of LLM-as-a-Judge behaviors helping bridge the gap between model understanding and human-centered risk understanding. In our paper, we make the following contributions:
\begin{enumerate}
    \item We propose a stakeholder driven policy explanation pipeline to observe the policy conflicts that emerge with different stakeholders using rule-based explanations. 
    \item We identify and explain how stakeholder conflicts emerge on three real-world AI usecases to demonstrate the practicality of the framework.
    \item We propose a tool for visualizing and interpreting these emerged conflicts. 
\end{enumerate}


\section{Literature Review}


\subsection{Persona in LLMs}

Assigning distinct personas to LLMs has emerged as a method to elicit different behaviours across tasks, showing that outputs vary under different assumed identities. For example, PersonaLLM \cite{jiang2024personallm} showed that GPT-3.5 and GPT-4, when assigned Big Five personality \cite{de2000big}, produce writings that aligned with those personas. Similarly \citet{hu2024quantifying}'s work on the effect of persona in LLM simulations show that incorporating persona (e.g. demographic) variables via prompting in LLMs provides modest but statistically significant improvements. More recently, \citet{dash2025persona} assigned 8 personas across 4 political and socio-demographic attributes to show that personas induce motivated reasoning in LLMs. These works indicate that inducing personas in LLMs alters LLM behaviour and can be a useful method to study outputs where grounding tasks on different personas can give us useful information about how personas are affected in those tasks.

\subsection{Prompt Robustness}

 Relying on a single prompt is brittle \cite{li2025humans} but aggregating results from meaning preserving paraphrases can lead to better claims. \citet{mizrahi2024state} show that averaging over an instruction paraphrase set (rather than choosing one prompt) yields more stable scores. \citet{wahle-etal-2024-paraphrase} showed that different paraphrase types (lexical, syntactic, morphological) elicit divergent behaviors and \citet{ceron2024beyond} argued that robustness to paraphrasing should be a standard reliability check alongside other perturbations. \citet{meier-etal-2025-towards} showed that distinct paraphrase types elicit measurably different behaviors even when semantics is held constant. This is particularly important when working with real world usecases. Imagine a scenario where an LLM is tasked to help diagnose a patient. Even if the LLMs gives results with highest predictions, if its results vary on different runs, a doctor can not rely on the evaluations of the LLM. Hence, prompt robustness is an important step in LLM evaluation to ensure the robustness of the output prediction.

\subsection{AI Risk Taxonomies}

Risk assessment is an integral step of AI system deployment. While several risk taxonomies have been proposed to map out the landscape of existing risks, the research on risk assessment is still quite fragmented. \citet{weidinger2022taxonomy} initially developed a comprehensive taxonomy of LLM risks, categorizing 21 distinct risk types ranging from discrimination and misinformation to malicious use and environmental harms into 6 broad categories. Subsequently, \citet{slattery2024ai} proposed  the AI Risk Repository, which looked at 777 risk statements from 43 prior frameworks and categorized them into a hierarchical structure into causal and domain-specific risks. Complimentary to this work, there has been an emergence in governance-oriented taxonomies to operationalize AI risk management like NIST \cite{nist}, OWASP \cite{owasp} and  IBM AI Risk Atlas \cite{bagehorn2025ai}. These frameworks have attempted to structure how AI risks are identified, categorized, and mitigated across organizational and model-governance dimensions to identify, measure, and mitigate technical and systemic risks.


\section{Methodology}

Our goal is to take risk prediction and stakeholder-grounded explanation of these risks (what we call \textit{policies}) to analyze how risks and their interpretations vary across different stakeholders in different AI use cases. We start with synthetic dataset of three real world use cases and their associated stakeholders to ground risk predictions in the stakeholders. Next, we use the IBM AI Risk Atlas Nexus for identifying potential risks within each stakeholder's context. Finally, we use the GloVE explanation framework to get stakeholder-specific policies that capture how different stakeholders reason about the same risks. The following sections describe each of these components in detail.

\subsection{Dataset Construction}

We introduce stakeholders as personas - which are intended to represent real world actors that are part of the system. Having these stakeholders serves two purposes: a) contextualize our pipeline such that predicted risk explanations are grounded in the role of the stakeholder in the usecase and b) for us to compare and analyse the differences in explanations of the diverse stakeholders for same usecase.  In this paper, we look at 3 unique \textit{base usecases} synthetically generated for different domains 1) AI medical diagnosis assistant that determines if someone needs surgery 2) Autonomous vehicle system that determines if passengers reach destination safely, and 3) AI fraud detection that determines if customer transactions get blocked. For each usecase, we then create stakeholder grounded usecases. Take for example a usecase ``AI medical diagnosis assistant that determines if someone needs surgery'' with following stakeholders: \textit{Surgeons, Primary Care Physicians, Radiologists, Patients requiring surgery, Patients with chronic conditions, Patients with acute injuries, Family members, Nurses and Healthcare administrators}. An example stakeholder grounded usecase can be ``\textit{surgeons using ai medical diagnosis assistant that determines if someone needs surgery}". We used a structured approach to generate this synthetic dataset as described below:

\begin{enumerate}
    \item We start with a list of base usecases for which we want to analyse risks.
    \item We then use gpt-4o to generate list of stakeholders (users and subjects) by asking top 3 high stake users, AI impacted subjects and secondary impacted subjects each. For consistency, we provide the definition of stakeholder from CSIRO responsible AI (RAI) catalogue\footnote{\url{https://research.csiro.au/ss/science/projects/responsible-ai-pattern-catalogue/}} in our prompt.
    \item For each stakeholder for each usecase, we generate a stakeholder specific usecase by substituting these stakeholders into base usecase. If the stakeholder is a user, we use the format: ``[STAKEHOLDER] using [base usecase]'' and if the stakeholder is a subject, we use the format ``[base usecase] that impacts [STAKEHOLDER]''.
    \item Finally to ensure the robustness in our experiments, inspired by \citet{meier-etal-2025-towards}, we generate paraphrases for these stakeholder specific usecase without changing their meaning. We use 6 different linguistic transformations a) addition/deletion b) semantic change c) same polarity substitution d) punctuation change e) change of order f) spelling change. Some examples of these paraphrases can be found in ~\cref{tab:linguistic_transformations}. 

Detailed prompt for generating this stakeholder specific usecases is provided in the Appendix in Listing~\ref{lst:stakeholder-paraphrase} and Table~\ref{tab:paraphrase_schema}. The goal is to take usecases grounded in these stakeholders and then use policy explanation pipeline on them to see how explanations differ and conflicts emerge. 
\end{enumerate}

\begin{table}[t]
\centering
\small
\caption{Examples of different linguistic transformations of a base stakeholder usecase \textit{surgeons using ai medical diagnosis assistant that determines if someone needs surgery.}}
\label{tab:linguistic_transformations}
\begin{tabular}{p{2.5cm} p{5.0cm}}
\toprule
\textbf{Transformation Type} & \textbf{Example} \\
\midrule
Addition / Deletion & surgeons are using an ai medical diagnosis assistant which determines whether a person requires surgery \\

Semantic Change & surgeons are utilizing an ai medical diagnosis tool which assesses whether surgery is necessary. \\

Same Polarity Substitution & surgeons utilizing ai healthcare diagnostic tool that determines if someone needs surgery \\

Punctuation Change & surgeons are using an ai medical diagnosis assistant that determines if someone needs surgery. \\

Change of Order & using an ai medical diagnosis assistant, surgeons determine if someone needs surgery \\

Spelling Change & surgeons using ai medical diagnosis assistant that determines if someone needs surgery. \\
\bottomrule
\end{tabular}
\end{table} 


\subsection{Risk Assessment}
\label{sec:risk-prediction}
We use IBM AI Risk Atlas \cite{bagehorn2025ai}, a comprehensive taxonomy of governance-related risks, as a taxonomy of risks for the model to predict from, along with Risk Atlas Nexus (RAN) \cite{bagehorn2025ai}, a risk assessment tool that uses Large Language Models (LLMs) to infer risks based on any given taxonomy, as shown in our pipleine in Figure~\ref{fig:pipeline}. It is worth noting that Risk Atlas Nexus builds on the IBM AI Risk Atlas but our method is agnostic to the risk assessment tool and taxonomy used and hence can be used with any taxonomy and risk prediction framework suited for the task. 

In our setup, as a first step, we use RAN to get risk predictions. However since our dataset comprises of paraphrase for each stakeholder per usecase, we start with taking the union of all risks inferred (from RAN) across all stakeholders as the complete set of possible risks for that use case. Next, for each stakeholder, we look at the paraphrased prompts that predicted at least one risk prediction. Finally, the risk is retained for a stakeholder if it is predicted consistently across all such paraphrases of that stakeholder. We classify all these risk-type as ``risk" and  all the risks types that are not retained are classified as ``not-a-risk". This approach ensures that the final risk set for each stakeholder reflects only stable, paraphrase-invariant predictions.  Formally, let $\mathcal{S}_u$ denote the set of stakeholders for a use case $u$, and let $\mathcal{P}_{u,s} = \{ p_{u,s,1}, p_{u,s,2}, \dots, p_{u,s,n_s} \}$ 
be the set of paraphrased prompts for stakeholder $s \in \mathcal{S}_u$. For each paraphrase $p_{u,s,i}$, the model outputs a set of predicted risks  $\mathcal{R}_{u,s,i} \subseteq \mathcal{R}_{\text{atlas}}$, 
where $\mathcal{R}_{\text{atlas}}$ is all the risks predicted for usecase based on the taxonomy of IBM AI Risk Atlas.

We first define the full set of risks identified for the use case as:
\begin{equation}
\mathcal{R}_{u} = 
\bigcup_{s \in \mathcal{S}_u} 
\bigcup_{i=1}^{n_s} 
\mathcal{R}_{u,s,i}.
\label{eq:r_usecase_union}
\end{equation}

We then retain only paraphrases that yield at least one predicted risk:
\begin{equation}
\mathcal{P}'_{u,s} = 
\{ p_{u,s,i} \in \mathcal{P}_{u,s} 
\mid |\mathcal{R}_{u,s,i}| > 0 \}.
\label{eq:valid_prompts}
\end{equation}

The final, consistent risk set for stakeholder $s$ is obtained by intersecting predictions across its valid paraphrases:
\begin{equation}
\mathcal{R}^*_{u,s} = 
\bigcap_{p_{u,s,i} \in \mathcal{P}'_{u,s}} 
\mathcal{R}_{u,s,i}.
\label{eq:consistent_stakeholder_risks}
\end{equation}

Thus, a risk $r \in \mathcal{R}_{\text{usecase}}$ is associated with stakeholder $s$
if and only if it appears in all paraphrases of $s$ that produced any prediction \footnote{We adopt a strict consensus (100\% intersection) rather than majority vote to avoid brittle, prompt-specific risks. However, this is a design choice and users can make it more flexible depending on their requirements.}.

\subsection{Explaining Stakeholder Conflicts}

After we compile a risk profile for each of the stakeholders, we can identify risks on which they agree and disagree in a given use case. To explain the differences in risk assessments across stakeholders, we utilize GloVE explanation pipeline. GloVE is a global explanation pipeline that extracts rule-based explanations using LLM-as-a-Judge. In this work, we use GloVE to explain risk assessment decisions by Risk Atlas Nexus from the perspective of individual stakeholders. For a given use case, for each stakeholder and each risk, we use GloVE to generate a set of rules using IF and DESPITE cluases that explain why the risk assessment might be relevant for that stakeholder. Formally,

\begin{equation}
\begin{aligned}
r_i \;\vdash_s\;&\;
\text{IF stakeholder-specific supporting concepts apply,} \\
&\text{DESPITE other contrasting factors.}
\end{aligned}
\label{eq:sglove_rule_structure_simple}
\end{equation}

where  $r_i$ is the risk identified within use case $u$ , $s$ is the stakeholder $s \in \mathcal{S}_u$ 



\section{Experiments and Results}

\begin{figure}[t]
  \centering
  \includegraphics[width=\linewidth]{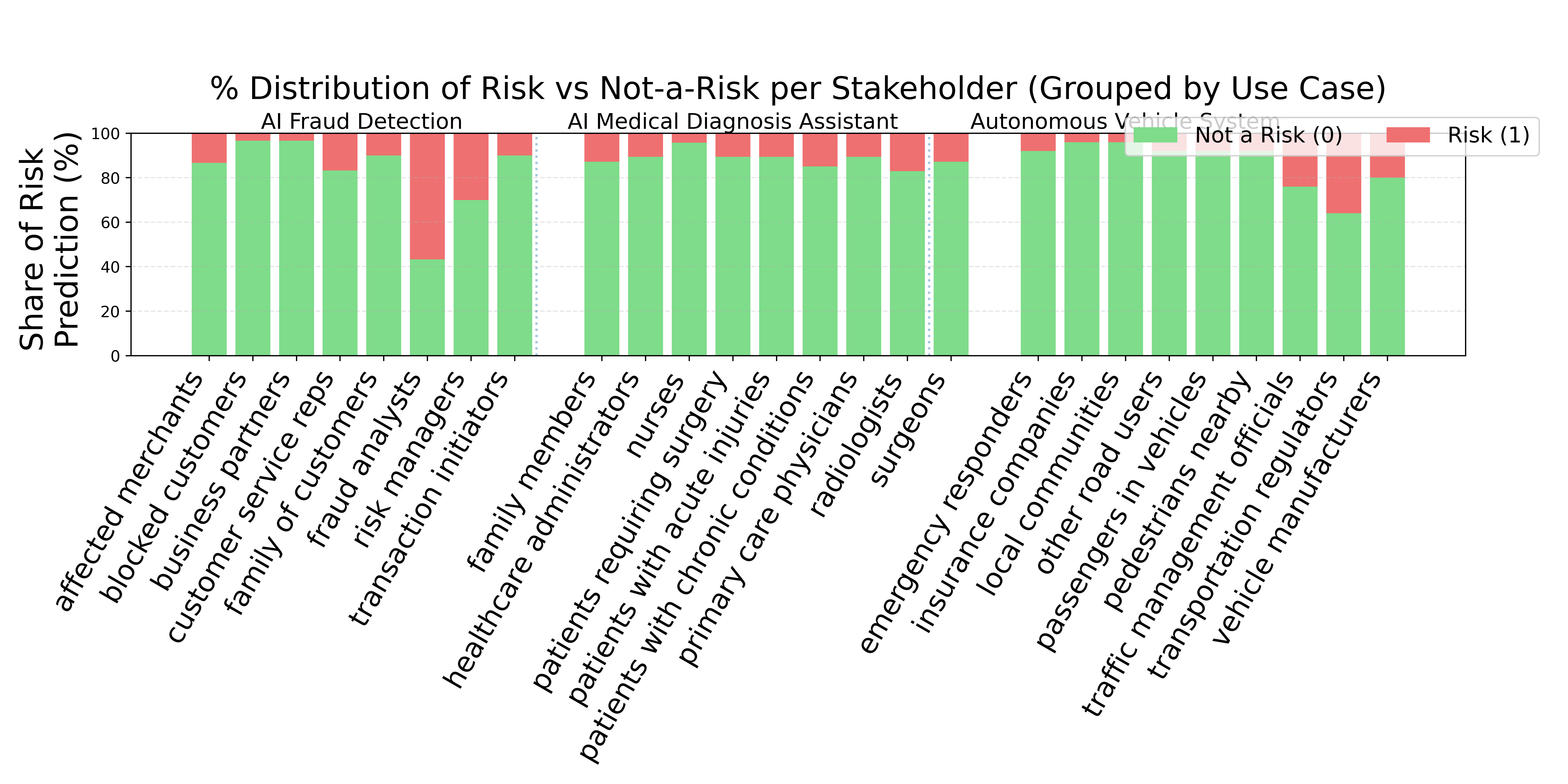}
  \caption{Risk assessment label distribution for all three usecase}
  \label{fig:medical-diagnosis}

\end{figure}





\begin{figure*}
  \includegraphics[width=1\linewidth]{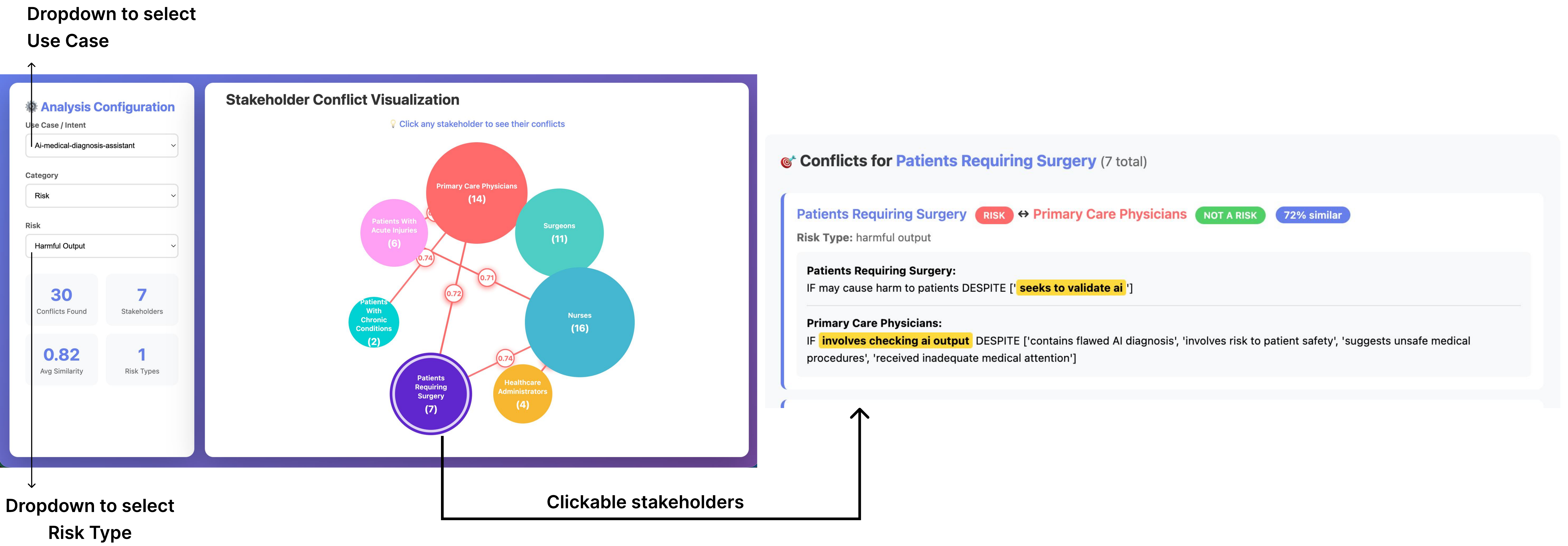}
    \caption{Stakeholder conflict visualizations for the \textit{AI Medical Diagnosis Assistant} use case. Each node represents a stakeholder, and edges indicate relationships based on overlapping or conflicting risk perceptions.}
    \label{fig:conflict-comparison}
\end{figure*}

\subsection{Stakeholder Risk Perceptions}
We first look at how stakeholders perceive risk within each use case. Figure~\ref{fig:medical-diagnosis} shows the distribution of number of predicted risks across stakeholders for all three usecases. Each stacked bar represents the share of risks labeled as \textit{\colorbox{red!30}{risk}} (label = 1) and \textit{\colorbox{green!30}{not-a-risk}} (label = 0). It is to note that we initially had 9 LLM generated stakeholder "AI Fraud detection`` usecase, however, one of the stakeholder produced 0 risks across all its paraphrases and was discarded from the results. Hence, we report results for the 8 remaining stakeholders for this usecase. The results show that risk perceptions vary by stakeholder role and context. It is worth noting  that design choice of choosing risk after stakeholder paraphrasing was very rigid as discussed before hence number of ``risk" classifications are low as compared to ``non-risks" that can be made more flexible to allow as per usecase. Overall, the results indicate substantial variation in how risks are perceived across stakeholders and use cases. For most stakeholders, the majority of predictions were classified as not-a-risk, reflecting the conservative nature of our strict consensus rule for retaining risks across paraphrases. However, some stakeholders e.g. fraud analysts in the fraud detection domain, family members \& patients in the medical diagnosis use case, and transportation regulators in the autonomous vehicle context show higher proportions of risk-labeled predictions. These patterns suggest that certain stakeholders are directly exposed to or affected by the AI system’s decisions. These variation support our core hypothesis that stakeholder roles and contexts significantly influence risk perception, underscoring the importance of stakeholder-grounded approaches for capturing the diversity of concerns in AI governance.

\subsection{Measuring Risk Conflicts}

Our experiments focus on using LLM-as-a-Judge to explain policies grounded in stakeholder-specific reasoning across diverse use cases. However, different stakeholders within the same usecase may often disagree on whether a given situation constitutes a risk or not. When this happens, we call this a \textit{conflict}. Formally, if $\mathcal{S}_u$ is a set of stakeholders and $\mathcal{R}_u$ is the risks for usecase $u$ such that risk $r_i\in\mathcal{R}_u$, $y_{u,s,i}\in\{0,1\}$ denotes the LLM-as-a-Judge label (1 = risk, 0 = not-a-risk) for stakeholder $s\in\mathcal{S}_u$ then conflict measure $\kappa_u(r_i)$ can be defined as:

\begin{equation}
\kappa_u(r_i)
=
\begin{cases}
1, & \text{if }\exists\, s_1,s_2:\, y_{u,s_1,i}\neq y_{u,s_2,i},\\[4pt]
0, & \text{otherwise.}
\end{cases}
\label{eq:conflict_indicator_case}
\end{equation}

where $\kappa_u(r_i)$ is an indicator that equals $1$ (hence conflict) if any two stakeholders in use case $u$ disagree on whether risk $r_i$ constitutes a risk, and $0$ otherwise. Table~\ref{tab:conflict-stats} shows the conflict rate statistics for all three usecases. We observe that stakeholders in usecase \textit{AI Fraud Detection} have most conflicts, followed by usecase \textit{Autonomous Vehicle} and \textit{AI Medical Diagnosis}.

\begin{table}[t]
\caption{Conflict statistics across the 3 chosen AI use cases}
\begin{adjustbox}{width=\linewidth}
\begin{tabular}{c|cccc} \toprule
AI Use Case          & Stakeholders & Risks & Conflicts & Average Conflict Rate \\ \midrule
AI Fraud Detection   & 8             & 30   &    20       &       66 \%                \\
AI Medical Diagnosis & 9             & 47   &    10       &       21.27 \%                \\
Autonomous Vehicle   & 9             & 25   &    14      &        56 \%            
\end{tabular}
\label{tab:conflict-stats}
\end{adjustbox}
\end{table}

\section{Policy Conflicts and Proposed Visualization}

An interesting subset of analyzing these conflicts is to see if the stakeholders are opposing due to similar perspectives for the same risk but from opposing directions.  For example, one stakeholder may state, ``We do not view this as a risk if the decision is always reviewed by a human,'' whereas another stakeholder within same usecase may claim,``We still consider this a risk despite human oversight.''. Hence, while their assessments conflict, their reasoning reflects similar considerations approached from opposing perspectives. To systematically understand such explanations in disagreements, we look at the explanations and analyze the \textsc{IF} and \textsc{DESPITE} clauses in the stakeholder-specific explanations. When a conflict arises  between two stakeholder,  then intuitively, we can look  \textsc{IF} justification and \textsc{DESPITE} justification for the same risk of different stakeholder and see if they are using \textit{similar} justifications but in opposite directions. 

In such cases, we define a conflict score between two stakeholders $s_1, s_2 \in \mathcal{S}_u$ for a given risk $r_i$ as:
\begin{equation}
\begin{aligned}
C_{u}(s_1, s_2, r_i) 
&= \max\Big(
\text{sim}(I_{u,s_1,i}, D_{u,s_2,i}), \\
&\qquad\;\;\text{sim}(I_{u,s_2,i}, D_{u,s_1,i})
\Big),
\end{aligned}
\label{eq:conflict_score}
\end{equation}

where $\text{sim}(\cdot,\cdot)$ measures the semantic similarity between two textual clauses (e.g., cosine similarity of sentence embeddings). A high $C_{u}(s_1, s_2, r_i)$ indicates that two stakeholders are  reasoning about the same underlying concept but from opposing directions: one as a reason to support the view (\textsc{IF}), the other framing it as an opposing reason (\textsc{DESPITE}). In our experiments, we use all-MiniLM-L6-v2 model to compute semantic similarity between stakeholders’ IF and DESPITE statements and identify conceptual conflicts.

To visualize these conflicts we propose an interactive visualization as seen in Figure~\ref{fig:conflict-comparison}. Stakeholders are color-coded bubbles, with their size reflecting the number of conflicts they're involved in. Conflicts are represented by connecting lines between stakeholder pairs. Users can filter by use case and specific risk types (e.g. harmful output), then click on any stakeholder to view detailed conflict information including full rule explanations with highlighted matching clauses that cause the disagreement. 
In Figure~\ref{fig:conflict-comparison}, we show our proposed method to visualize the stakeholder conflicts identified for the usecase \textit{AI Medical Diagnosis Assistant} with respect to risk type ``harmful output". We believe such a tool would be valuable to visualize conflicts and understand what led to those conflicts.



\section{Discussion}
In this paper, we introduced a stakeholder-grounded framework for AI risk assessment that uses LLM-as-a-Judge to predict and explain risks through stakeholder-specific usecases. Our findings highlight that stakeholder perspectives play a central role in shaping how risks are perceived and explained within AI systems. By grounding risk predictions and explanations in stakeholder contexts, our framework shows variations in risk assessments that traditional, stakeholder-agnostic evaluations overlook. We also propose an interactive visualization to enhance transparency of the conflict reasoning that emerge. Beyond interpretability, our stakeholder-aware explanations paves a way for more transparent and auditable LLM-based evaluations.Together, these contributions point toward a future where stakeholder-aware, explainable evaluations can form the backbone of trustworthy AI governance.

\section{Limitations}

The current approach relies on synthetic, LLM-generated stakeholders, which although provides for scalability, may not fully capture the complexity and unpredictability of real-world scenarios. Our framework also focuses on a single risk taxonomy and a risk assessment tool whose performance is dependent on the robustness of their underlying components. Finally, our framework gives binary outcomes as either ``risk" or ``not-a-risk" which can be improved to cover more range of risk categories  e.g., low, medium, high, critical (e.g. as proposed in the EU AI Act). Future work could integrate real stakeholder feedback, multiple taxonomies, and graded risk levels to improve the granularity of stakeholder-aware evaluations.

\section{Acknowledgments}
This work was funded by the EU Horizon project ELIAS (No. 101120237). Views and opinions expressed are those of the author(s) only and do not necessarily reflect those of the European Union or The European Research Executive Agency.

\bibliography{aaai2026}

\appendix

\section{Appendix}
\label{sec:appendix}

\label{sec:data-generation-details}

\lstdefinestyle{mystyle}{
    backgroundcolor=\color{backcolour},   
    commentstyle=\color{codegreen},
    keywordstyle=\color{magenta},
    numberstyle=\tiny\color{codegray},
    stringstyle=\color{codepurple},
    basicstyle=\ttfamily\tiny,
    breakatwhitespace=true,         
    breaklines=true,
    captionpos=b,                    
    keepspaces=true,                 
    numbers=none,                    
    numbersep=5pt,                  
    showspaces=false,                
    showstringspaces=false,
    showtabs=false,                  
    tabsize=2,
    xleftmargin=2em,
    xrightmargin=2em
}

\lstset{style=mystyle}

\begin{table*}[t]
\centering
\caption{Structured overview of all paraphrase types used in the stakeholder-specific dataset generation, showing their definitions, example transformations, and corresponding prompt templates. The base prompt that uses these details can be seen in Listing~\ref{lst:stakeholder-paraphrase}.}
\label{tab:paraphrase_schema}
\tiny  
\setlength{\tabcolsep}{3pt}
\renewcommand{\arraystretch}{1.15}

\begin{tabular}{p{0.12\linewidth} p{0.15\linewidth} p{0.24\linewidth} p{0.40\linewidth}}
\toprule
\textbf{Paraphrase Type} & \textbf{Definition} & \textbf{Example (Input –Output)} & \textbf{CoT Reasoning Example } \\
\midrule

Addition / Deletion &
Addition/Deletion consists of all additions/deletions of lexical and functional units. &
\textit{Input:} “Revenue in the first quarter of the year dropped by 15 percent from the same period a year earlier.”  
\newline
\textit{Output:} “Revenue in the first quarter of the year only dropped 15 percent from the same period a year earlier.” &
The task is about paraphrasing using adding/deletion such that meaning of the input sentence is preserved. So, I can add the word "only" before "dropped" to slightly emphasize the extent of decline without altering the meaning. Then I can also remove the word "by" after "dropped," since it is optional for sentence and does not affect the meaning. \\

\hline
Semantic-based Change &
Semantics-based changes involve a different lexicalization of the same content units, often affecting multiple words. &
\textit{Input:} “WalMart said it would verify the employment status of all its million-plus domestic workers to ensure they were legally employed.”  
\newline
\textit{Output:} “WalMart announced that it would verify the legal employment status of all its million-plus domestic workers.” &
The task is about paraphrasing using semantics-based changes which can involve re-expressing the same content units using different lexicalizations that often affect multiple words together. In this case, I can change the reporting phrase "said it would" into "announced that it would," which is not a single-word substitution but a different way of expressing the act of communication. I can also transform the purpose clause "to ensure they were legally employed" into the adjectival phrase "legal employment status." This change spans multiple lexical units and shows how the meaning is preserved but expressed differently. The part "all its million-plus domestic workers" was kept intact to preserve the scope of the content. Together, these edits align with semantics-based changes because they alter how the meaning is lexicalized rather than simply adding or deleting words. \\
\hline

Same Polarity Substitution &
Changing one lexical unit for another with approximately the same meaning, such as synonymy or general/specific alternation. &
\textit{Input:} “They had published an advertisement on the Internet on June 10.”  
\newline
\textit{Output:} “They had posted an advertisement on the Internet on June 10.” &
The task is to paraphrase using same-polarity substitution, which means swapping a lexical unit with another that carries approximately the same meaning—typically via (for e.g.) synonymy without altering the proposition's content or sentiment. Here, I can replace "published" with "posted" because, in an online context, both verbs denote making material publicly available, preserving the event type and polarity. I will keep all the words unchanged to maintain participants, setting, and timeline. This is a like-for-like predicate substitution aligning with the definition. \\
\hline

Punctuation Change &
Any change in punctuation or sentence formatting without altering lexical units. &
\textit{Input:} “Trading in Loral was halted yesterday. The shares closed on Monday at \$3.01.”  
\newline
\textit{Output:} “Trading in Loral was halted yesterday; the shares closed on Monday at \$3.01.” &
The task is to paraphrase using punctuation and format changes, which means modifying how the sentence is punctuated or structured without altering the lexical units themselves. In this case, I can merge the two related sentences into one by replacing the period after "yesterday" with a semicolon. This adjustment will signal a close connection between the two clauses while keeping all the words unchanged. The meaning and content will also remain identical. This aligns with the definition since the change is purely in punctuation, not in vocabulary." \\
\hline

Change of Order &
Reordering words, phrases, or clauses while maintaining the same meaning. &
\textit{Input:} “The processors were announced in San Jose at the Intel Developer Forum.”  
\newline
\textit{Output:} “The processors were announced at the Intel Developer Forum in San Jose.” &
The task is to paraphrase using change of order, which means re-arranging the position of words, phrases, or clauses while keeping the meaning intact. In this case, I can shift the location of the phrase "in San Jose" from before "at the Intel Developer Forum" to after it. This reordering does not alter the entities involved, the event (the processors being announced), or the location information, but it changes how the sentence is structured. Since only the order of the phrases is modified and no lexical items are added or removed, this matches the definition of change of order." \\
\hline

Spelling Change &
Altering the spelling or written format (e.g., case changes, abbreviations, or digit/letter alternations) while preserving meaning. &
\textit{Input:} “It said the damage to the wing provided a pathway for hot gasses to penetrate the ship's thermal armor during Columbia's ill-fated reentry.”  
\newline
\textit{Output:} “It stated that the damage to the wing provided a pathway for hot gases to penetrate the ship's thermal armour during Columbia's ill-fated re-entry.” &
The task is to paraphrase using spelling and format changes, which involve altering the orthography or written form of lexical units without changing their meaning. In this case, I can replace the American English spelling "gasses" with the more standard form "gases," and "armor" with the British English variant "armour." I can also reformat "reentry" as "re-entry" by adding a hyphen. These changes do not affect the semantics of the sentence but strictly concern spelling and format conventions, which aligns directly with the definition." \\
\bottomrule
\end{tabular}
\end{table*}

\begin{figure*}[!b]
\centering
\begin{lstlisting}[
  label=lst:stakeholder-paraphrase,
  caption={Base format of the paraphrase prompt.},
  basicstyle=\ttfamily\tiny,
  breaklines=true,
  columns=fullflexible,
  frame=single,
  xleftmargin=0pt, xrightmargin=0pt,
  aboveskip=2pt, belowskip=2pt
]
"paraphrase_1": {
    "type": <paraphrase type>,
    "definition": <definition of the paraphrase type>,
    "example": [
      "Input: <input usecase>,
       CoT Reasoning: <example reasoning>,
       Output: <example output>"
    ],
    "prompt": [
      "In this task you will be given a definition of an alteration and an input sentence...",
      "Output the altered sentence at the end with 'Output:' in front.",
      "",
      "Alteration: {definition}",
      "Example: {example}",
      "Input: {usecase}"
    ]
  }
\end{lstlisting}
\end{figure*}

\end{document}